\pdfoutput=1

\documentclass[11pt]{article}
\usepackage{nodalida2023}
\usepackage{times}
\usepackage{url}
\usepackage{hyperref}
\usepackage{latexsym}

\usepackage[T1]{fontenc}

\usepackage[utf8]{inputenc}

\usepackage{microtype}

\usepackage{graphicx}

\usepackage{longtable}

\usepackage{booktabs}

\aclfinalcopy 

\newcommand\pkgurl{\url{https://github.com/saattrupdan/ScandEval}}
\newcommand\leaderboardsourceurl{\url{https://github.com/ScandEval/scandeval.github.io}}
\newcommand\leaderboardurl{\url{https://scandeval.github.io}}
\newcommand\hfurl{\url{https://huggingface.co/ScandEval}}
\newcommand\scandiqahfurl{\url{https://huggingface.co/datasets/alexandrainst/scandi-qa}}
\newcommand\scandiqarepourl{\url{https://github.com/alexandrainst/ScandiQA}}

\title{ScandEval: A Benchmark for Scandinavian Natural Language Processing}

\author{
  Dan Saattrup Nielsen \\
  The Alexandra Institute \\
  Copenhagen, Denmark \\
  \texttt{dan.nielsen@alexandra.dk}
}

\date{}

\begin{document}
\maketitle
\begin{abstract}
    This paper introduces a Scandinavian benchmarking platform, \texttt{ScandEval},
    which can benchmark any pretrained model on four different tasks in the
    Scandinavian languages. The datasets used in two of the tasks, linguistic
    acceptability and question answering, are new. We develop and release a Python
    package and command-line interface, \texttt{scandeval}, which can benchmark any
    model that has been uploaded to the Hugging Face Hub, with reproducible results.
    Using this package, we benchmark more than 100 Scandinavian or multilingual models
    and present the results of these in an interactive online
    leaderboard\footnote{\leaderboardurl}, as well as provide an analysis of the
    results. The analysis shows that there is substantial cross-lingual transfer among
    the Mainland Scandinavian languages (Danish, Swedish and Norwegian), with limited
    cross-lingual transfer between the group of Mainland Scandinavian languages and the
    group of Insular Scandinavian languages (Icelandic and Faroese). The benchmarking
    results also show that the investment in language technology in Norway, Sweden and
    Denmark has led to language models that outperform massively multilingual models
    such as XLM-RoBERTa and mDeBERTaV3. We release the source code for both the
    package\footnote{\pkgurl} and leaderboard\footnote{\leaderboardsourceurl}.
\end{abstract}

\section{Introduction}
\label{sec:introduction}
In recent years, there has been a significant increase in the number of monolingual
language models in the Scandinavian languages \cite{mollerhoj2020nordicbert,
    hojmarkbertelsen2021lctra, sarnikowski2021danishtransformers, enevoldsen2021dacy,
    abdaoui-etal-2020-load, kummervold-etal-2021-operationalizing,
malmsten2020swedishbert, snaebjarnarson2023transfer}, to the
extent that it becomes difficult both for the practioner to choose the best model for
the task at hand, as well as for language researchers to ensure that their research
efforts are indeed improving upon past work.

Aside from the increasing number of models, \citet{sahlgren-etal-2021-basically} also
emphasises that a joint Scandinavian language model is probably a better strategy for
the Scandinavian countries, considering the similarity of their languages and culture.
Indeed, \citet{faarlund2019syntax} even argues that the Danish, Norwegian and Swedish
languages are so similar that they should be considered a single language.

The languages included in the term ``Scandinavian'' is debatable
\cite{oxford2021scandinavia}. Following the distinction between \textit{Mainland
Scandinavian} (Danish, Swedish and Norwegian) and \textit{Insular Scandinavian}
(Icelandic and Faroese) \cite{haugen1976scandinavian, faarlund2019syntax}, a
distinction based on mutual intelligibility and syntactical structure, we focus in this
work on the Mainland Scandinavian languages, while still allowing support for the
Insular Scandinavian languages. Aside from being a standard distinction, our choice is
also based on experiments on the cross-lingual transfer between these two groups, which
we present in Section~\ref{sec:crosslingualtransfer}. We will here use the term
``Scandinavian'' to mean the collection of all five languages, and use the
Mainland/Insular distinction when applicable.

To help facilitate progress in both improving upon the monolingual Scandinavian models
as well as the multilingual, we present \texttt{ScandEval}, a benchmark of Scandinavian
models, along with a Python package and Command-Line Interface (CLI), and an associated
online leaderboard. This leaderboard contains the results of language models
benchmarked on datasets within the Mainland Scandinavian languages, as described in
Section~\ref{sec:tasks}.

Recent studies \cite{khanuja-etal-2021-mergedistill, pires-etal-2019-multilingual,
lauscher-etal-2020-zero} have shown that multilingual models can outperform monolingual
models when the languages are sufficiently similar, and also that they are worse than
the monolingual models when the languages are too dissimilar. This shows that the
Scandinavian languages could have something to gain by creating ``local multilingual''
models, rather than using the massively multilingual models such as
\texttt{XLM-RoBERTa} \cite{conneau2020unsupervised}. Based on this, we test the
following hypotheses:
\begin{itemize}
    \item \textbf{Hypothesis 1:} There is a substantial cross-lingual transfer within
        the Mainland Scandinavian languages.
    \item \textbf{Hypothesis 2:} There is no notable cross-lingual transfer between
        the group of Mainland Scandinavian languages and the group of Insular
        Scandinavian languages.
\end{itemize}

To the best of our knowledge, this is the first benchmarking tool for any of the
Scandinavian languages, as well as the first online leaderboard containing scores from
such a tool. Our contributions are the following:
\begin{enumerate}
    \item We construct a new question answering dataset for the Mainland Scandinavian
        languages, dubbed \texttt{ScandiQA}.
    \item We construct a new linguistic acceptability dataset for all the Scandinavian
        languages, dubbed \texttt{ScaLA}.
    \item We develop a Python package and CLI, \texttt{scandeval}, which allows
        reproducible benchmarking of language models on Scandinavian language datasets.
    \item We uniformise all the datasets used in the benchmark, to enable consistent
        evaluation across languages and datasets. These uniformised datasets are also
        available on the Hugging Face Hub\footnote{\hfurl}.
    \item We benchmark all the Scandinavian and a selection of the multilingual
        language models on the Hugging Face Hub\footnote{\url{https://hf.co}} on the
        Mainland Scandinavian datasets in the benchmark, and present all the scores in
        an online leaderboard.
\end{enumerate}

\section{Related Work}
\label{sec:relatedwork}
There has been a number of (non-English) NLU benchmarks published in recent years
\cite{wang-etal-2018-glue, sarlin2020superglue, rybak-etal-2020-klej,
    ham-etal-2020-kornli, shavrina-etal-2020-russiansuperglue, wilie-etal-2020-indonlu,
    xiang-etal-2021-climp, koto-etal-2020-indolem, Safaya2022MukayeseTN,
Augustyniak2022ThisIT, Khashabi2020ParsiNLUAS, Ham2020KorNLIAK, Xu2020CLUEAC,
Dumitrescu2021LiRoBA}, with whom we share the same goal of advancing the state of NLP
in our respective languages. Within the Scandinavian languages specifically, the
SuperLim benchmark \cite{adesam2020swedishglue} is a Swedish NLU Benchmark featuring
several difficult tasks. Most of the datasets in the SuperLim benchmark only contain a
test set, however.

The \texttt{XGLUE} \cite{liang-etal-2020-xglue} dataset is another multilingual NLU
benchmark. That dataset is different from \texttt{ScandEval} in that all the training
data in \texttt{XGLUE} is in English, and that the majority of the test sets are not
available in any of the Scandinavian languages.

\citet{isbister2020not} present a Swedish similarity benchmark, achieved through
machine translating the \texttt{STS-B} dataset from the GLUE benchmark
\cite{wang-etal-2018-glue}. Aside from only dealing with a single task and a single
language, the quality of the dataset is worse than a gold-standard corpus as a result
of the translation, as the authors also point out.

\section{Methodology}
\label{sec:methodology}
This section describes our benchmarking methodology in detail, including both the setup
of the datasets, the evaluation procedure and the scoring of the models. We also
describe how we conduct the cross-lingual transfer experiments.

\subsection{Finetuning Setup}
\label{sec:finetuningsetup}
When finetuning, we enforce a learning rate of $2\cdot 10^{-5}$ with 100 warmup steps,
and a batch size of $32$. If there is not enough GPU memory to finetune the model with
this batch size, we halve it and double the amount of gradient accumulation, resulting
in the same effective batch size. This is repeated until the batches can fit in memory.

We impose a linear learning rate schedule with intercept after $10,000$ training steps
(with a training step consisting of 32 samples), and we adopt early stopping
\cite{plaut1986experiments} to stop the training procedure if the validation loss has
not decreased for 90 training steps. We use the \texttt{AdamW} optimiser
\cite{loshchilov2017decoupled} with first momentum $\beta_1=0.9$ and second momentum
$\beta_2=0.999$, and we optimise the cross-entropy loss throughout all tasks. Further,
random seeds are fixed throughout, to ensure reproducibility.

The finetuning itself uses the \texttt{transformers} package
\cite{wolf-etal-2020-transformers}. For the named entity recognition task we use the
\texttt{AutoModelForTokenClassification} class, which linearly projects the embedding
from the language model encoder for each token into the entity logits for that token.
For the classification tasks we use the \texttt{AutoModelForSequenceClassification}
class, which linearly projects the embedding from the language model encoder to each
document into the class logits for that document. Lastly, for the question answering
task we use the \texttt{AutoModelForQuestionAnswering} class, which linearly projects
the embedding from the language model encoder for each token, into the logits of the
start and end positions of the answer for that token.

\subsection{Bootstrapping Evaluation}
\label{sec:bootstrapping}
For each model and dataset, we repeat the following procedure 10 times, which generates
a score for each model and dataset combination: (a) Fix a random seed unique to the
given iteration; (b) Finetune the model on the training set; (c) Evaluate the model on
a bootstrapped (i.e., sampling with replacement) version of the test set. The
evaluation score is then the mean $\mu$ of these scores, along with a 95\% confidence
interval $I_{10}$, computed as

\begin{equation}
    \label{eq:confidenceinterval}
    I_N := \mu\pm\frac{1.96}{N-1}\sum_{i=1}^N\texttt{score}_i.
\end{equation}

The combination of varying the random seeds as well as using bootstrapped test datasets
ensures that we capture the noise coming from both the random initialisation of the
added layers to the model as well as the noise in the test set, resulting in a more
reliable confidence interval of the true mean for each model and dataset combination.

To aggregate these scores across all datasets, we firstly compute the
\textit{language-specific task scores} for each (model, language, task) triple, which
is the mean of the scores of the model on the tasks of the language.\footnote{This mean
    is only non-trivial for the Norwegian language for the named entity recognition
task and the linguistic acceptability task, as these tasks are available in both
Norwegian Bokmål and Norwegian Nynorsk.} From these language-specific scores we next
compute the \textit{language score} for each (model, language) pair as the mean of the
language-specific task scores across all the tasks. A final \texttt{ScandEval} score is
computed as the average of the language scores, to emphasise the training of
Scandinavian models rather than monolingual ones.

\subsection{Cross-lingual Transfer}
\label{sec:crosslingualtransfer}
To test Hypothesis 1 and 2, stated in Section~\ref{sec:introduction}, we introduce a
way to measure the ``joint cross-lingual transfer'' of a group of languages, by which we
mean an aggregate of the cross-lingual transfer between any two languages in the group.

To do this, we first introduce a control group of non-Scandinavian languages: English,
German, Dutch, Finnish, Russian and Arabic. By considering the combined set of
languages in the control group and the Scandinavian languages, we aim to find the best
split of these languages into two groups: a ScandEval benchmark group and a
non-benchmark group. The ``goodness'' of a split is measured by benchmarking a
``representative'' model from each language on datasets in each of the benchmark
languages and measuring the quality of the two-cluster clustering of these benchmarking
values.

As an example, if Danish and Swedish constitute the benchmark group and the rest of the
languages are in the non-benchmark group, we would benchmark the representative models
from each language on the Danish and Swedish part of ScandEval, and then compute the
F-statistic of the clustering $\{\{\texttt{da},\texttt{sv}\}, \{\texttt{no},
\texttt{is}, \texttt{fo}, \dots\}\}$ with these benchmarking values, computed as the
ratio of the between-group variance to the within-group variance.\footnote{Technically
    speaking, we get an F-statistic for each language in the benchmarking group, but we
just use the mean of these F-statistics.} We can then compare this F-statistic to the
F-statistic of the clustering where the benchmark group consists of Danish and
Norwegian, for instance.

As for picking a representative model for each language, we found pretrained language
models of roughly the same size on the Hugging Face Hub, each of which has been
pretrained on solely monolingual data. We note that no Faroese language model exists,
so for that language we do not include any model but still include Faroese benchmarking
datasets when Faroese is part of the benchmarking group. See the full list of models in
the appendix.

We can then restate our first hypothesis as the mainland Scandinavian languages are all
in the best-performing benchmark group, and our second hypothesis as the Insular
Scandinavian languages are not in the best-performing benchmark group.

\subsection{Uniform Benchmarking Datasets}
\label{sec:uniformdatasets}
As we are interested in comparing the performance of the models across languages, we
ensure that all the datasets used in the benchmark are of the same format and the same
size.

\begin{figure*}[!htb]
    \center
    \includegraphics[width=0.9\textwidth]{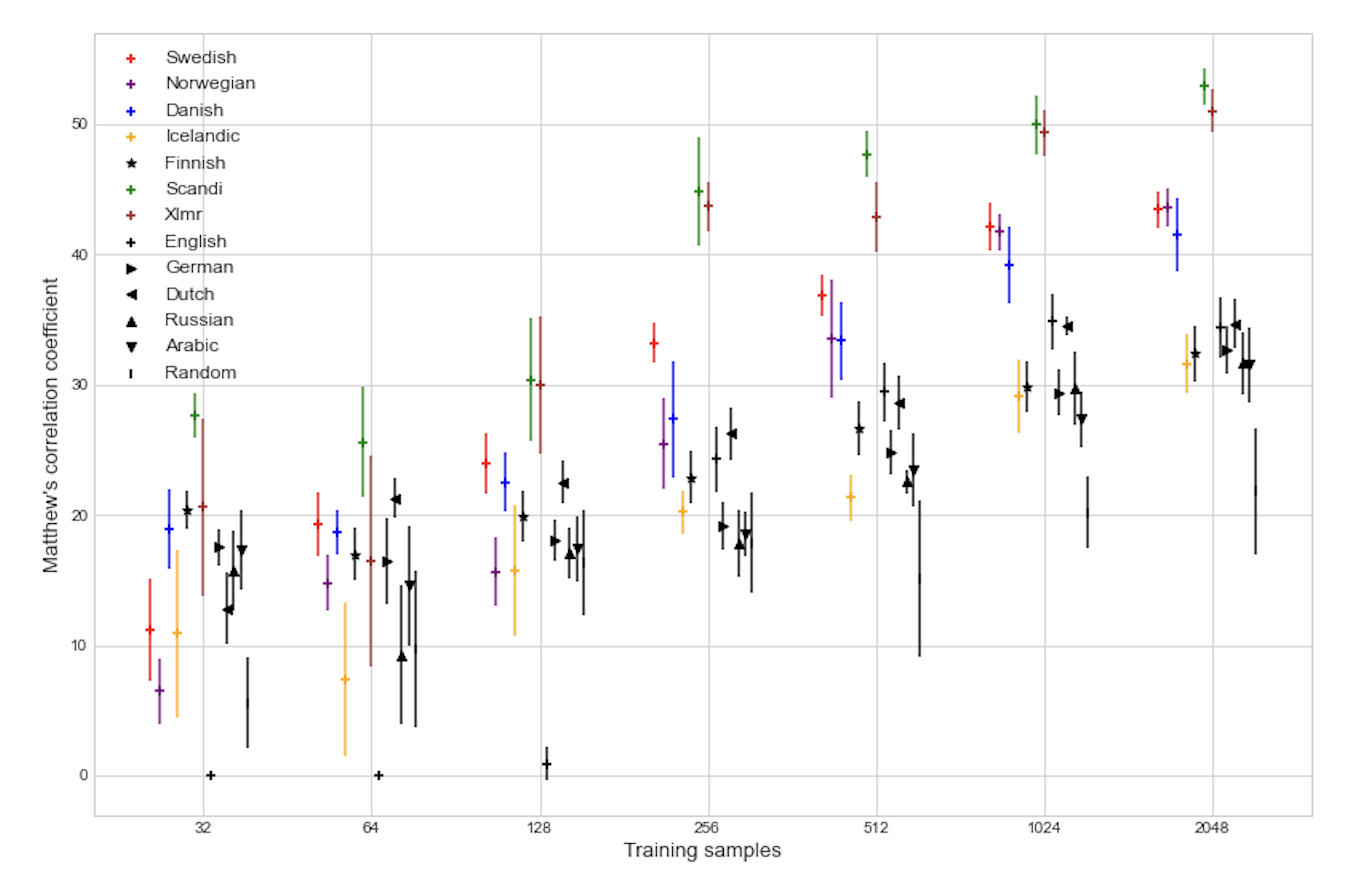}
    \caption{Plot showing the performance of different models on the
    \texttt{AngryTweets} dataset with varying number of training samples.}
    \label{fig:angrytweetstrainingsize1}
\end{figure*}

\begin{figure*}[!htb]
    \center
    \includegraphics[width=0.9\textwidth]{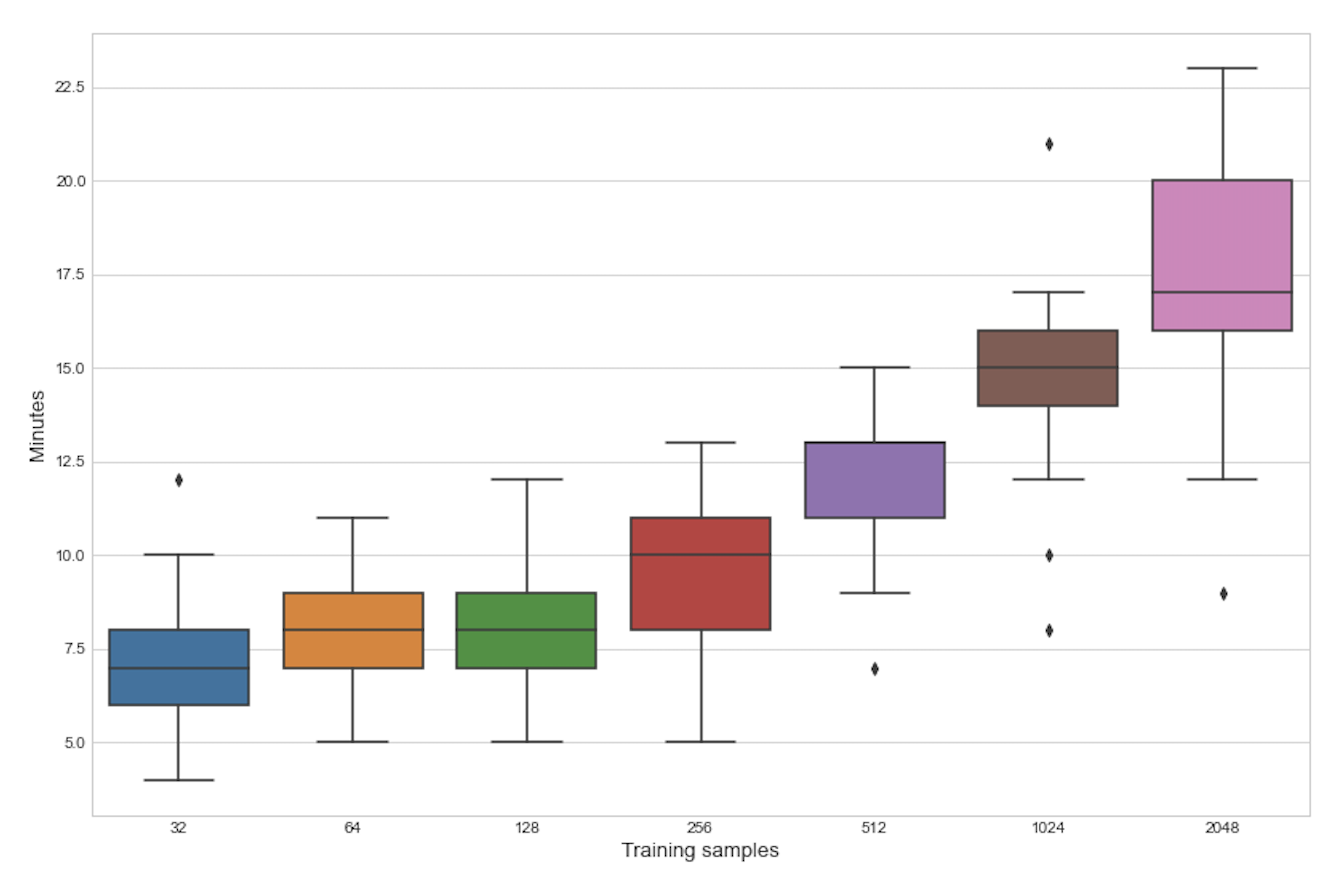}
    \caption{Boxplot showing the training time of the models on the
    \texttt{AngryTweets} dataset with varying number of training samples.}
    \label{fig:angrytweetstrainingsize2}
\end{figure*}

We aimed to choose a training data size that would be a balance between being able to
differentiate between the models and being able to benchmark the models in a reasonable
amount of time. We benchmarked the same models as in
Section~\ref{sec:crosslingualtransfer} on truncations of named entity recognition
datasets, sentiment classification datasets and linguistic acceptability datasets.
Based on these results we qualitatively found that using 1,024 training samples allowed
for both differentiation between the models and being able to benchmark the models in a
reasonable amount of time. Figures~\ref{fig:angrytweetstrainingsize1} and
\ref{fig:angrytweetstrainingsize2} show the trade-off between differentiation and
benchmarking speed, covering the \texttt{AngryTweets} dataset
\cite{pauli-etal-2021-danlp}. The remaining plots for the other datasets can be found
in the appendix.

Another benefit of using a small training dataset is that it emphasises the importance
of the pre-trained weights of the models, rather than the fine-tuning process. Further,
we wanted the test dataset to be as large as possible, to ensure more robust
evaluations of the models, which led to the choice of 2,048 test samples based on the
number of available samples in the smallest dataset. Lastly, the validation set was
chosen to be 256 samples, to allow for a reasonable evaluation during training, while
not being too time-consuming. All of these datasets with their splits are available on
the Hugging Face Hub.

\section{ScandEval Tasks}
\label{sec:tasks}
To properly evaluate the performance of a pretrained model, we ideally need to evaluate
it on many diverse tasks. Unfortunately, the Scandinavian languages do not have many
openly available datasets for many downstream tasks.

To address this, we construct two new Scandinavian datasets, \texttt{ScaLA} and
\texttt{ScandiQA}, being \textit{Linguistic Acceptability} (\texttt{LA}) and
\textit{Question Answering} (\texttt{QA}) datasets, respectively. These new tasks are
supplemented by existing benchmarking datasets within \textit{Named Entity Recognition}
(\texttt{NER}) and \textit{Sentiment Classification} (\texttt{SENT}). Aside from
downstream performance of these tasks, we also benchmark the inference speed of each
model. We describe all of these in more detail in the subsections below.

\subsection{Named Entity Recognition}
\label{sec:ner}
For the NER task we use the four classes used in CONLL
\cite{tjong-kim-sang-de-meulder-2003-introduction}: \texttt{PER}, \texttt{LOC},
\texttt{ORG} and \texttt{MISC}, corresponding to person names, locations, organisations
and miscellaneous entities.

Since this is a token classification task and that the language models usually use
different tokenisers, we have to ensure a uniform treatment of these as well. We
tokenise the documents using the pretrained tokeniser associated to the model that we
are benchmarking, and to ensure consistency of the evaluation we replace all but the
first token in each word with the empty entity \texttt{O}. For instance, if the word
``København'' with the \texttt{LOC} tag is tokenised as [``Køben'', ``havn''], then we
would assign the labels \texttt{LOC} and \texttt{O} to these tokens. This ensures that
we maintain the same number of (non-empty) labels per document.

In terms of evaluation metrics, we use the micro-average F1-score, which is standard
for NER. We also report a \textit{no-misc score}, which is the micro-average F1-score
after we replace the \texttt{MISC} class in the predictions and labels with the ``empty
label'' \texttt{O}. This \textit{no-misc score} is not used in any of the aggregated
scores and is purely used for comparison purposes on the individual datasets.

For Danish we use the \texttt{DaNE} dataset \cite{hvingelby-etal-2020-dane}, being a
NER tagged version of the Danish Dependency Treebank \cite{kromann2004danish}.
\texttt{DaNE} is already in the CONLL format, so we perform no preprocessing on the
data.

For Norwegian we use the Bokmål and Nynorsk \texttt{NorNE} datasets
\cite{jorgensen-etal-2020-norne}, also being NER tagged versions of the Norwegian
Dependency Treebanks \cite{ovrelid-hohle-2016-universal}. Aside from the \texttt{PER},
\texttt{LOC}, \texttt{ORG} and \texttt{MISC} tags, these also include
\texttt{GPE\_LOC}, \texttt{GPE\_ORG}, \texttt{PROD}, \texttt{DRV} and \texttt{EVT}
tags. We convert these to \texttt{LOC}, \texttt{ORG}, \texttt{MISC}, \texttt{MISC} and
\texttt{MISC}, respectively.

Lastly, Swedish does not have a NER tagged version of the corresponding dependency
treebank, but they instead have the \texttt{SUC3} dataset, a NER-enriched version of
the \textit{Stockholm-Umeå Corpus} \cite{gustafson2006manual}. This dataset does not
follow the CONLL format and is instead released in the XML format, with the
\texttt{<name>} XML tags containing the NER tags for the words they span over
\footnote{The \texttt{<ne>} XML tags are also NER tags, but these have been
automatically produced by \texttt{SpaCy} \cite{honnibal2020spacy} models and are thus
not gold standard.}. This dataset contains the NER tags \texttt{animal},
\texttt{event}, \texttt{inst}, \texttt{myth}, \texttt{other}, \texttt{person},
\texttt{place}, \texttt{product} and \texttt{work}. These were converted to
\texttt{MISC}, \texttt{MISC}, \texttt{ORG}, \texttt{MISC}, \texttt{MISC}, \texttt{PER},
\texttt{LOC}, \texttt{MISC} and \texttt{MISC}, respectively.

\subsection{Sentiment Classification}
\label{sec:sent}
We treat the sentiment classification task as a three-class classification task, with
the classes \texttt{positive}, \texttt{neutral} and \texttt{negative}. Evaluation of
the models is done using Matthew's Correlation Coefficient
\cite{matthews1975comparison} as the primary metric as well as reporting the
macro-average F1-score as a secondary metric. We choose to use Matthew's Correlation
Coefficient as the primary metric as it has been shown to be more reliable than the
macro-average F1-score \cite{chicco2020advantages}, while also being the standard
metric used in the \texttt{GLUE} \cite{wang-etal-2018-glue} and \texttt{SuperGLUE}
\cite{sarlin2020superglue} benchmarks.

For Danish we use the sentiment classification dataset \texttt{AngryTweets}
\cite{pauli-etal-2021-danlp}, which contains crowdsourced annotations of Danish tweets.
To comply with Twitter's Terms of Use we have fully anonymised the tweets by replacing
all user mentions with \texttt{@USER} and all links by \texttt{[LINK]}, as well as
shuffling the tweets.

For Norwegian we included the sentiment classification dataset \texttt{NoReC}
(Norwegian Review Corpus) \cite{velldal-etal-2018-norec}, which are based on scraped
reviews from Norwegian websites.

Lastly, for Swedish we use the sentiment classification dataset presented in
\citet{svensson2017sentiment}, which is based on reviews from the Swedish websites
\url{www.reco.se} and \url{se.trustpilot.com}. In analogy with \texttt{NoReC} we dub
this dataset the Swedish Review Corpus (\texttt{SweReC}).

\subsection{Linguistic Acceptability}
\label{sec:la}
Based on the inclusion of the CoLA (Corpus of Linguistic Acceptability) dataset
\cite{warstadt2019neural} in the GLUE benchmark \cite{wang-etal-2018-glue}, we
construct new linguistic acceptability datasets for the Scandinavian languages. This
task is often framed as a binary classification task, where the model is tasked with
predicting whether a given sentence is grammatically correct or not.

We dub our new datasets Scandinavian Linguistic Acceptability (\texttt{ScaLA}), which
we release for Danish, Norwegian Bokmål, Norwegian Nynorsk, Swedish, Icelandic and
Faroese. Each of these datasets consist of 1,024 training samples, 256 validation
samples and 2,048 test samples, in accordance with Section~\ref{sec:uniformdatasets}.
The \texttt{ScaLA} datasets are based on the Danish, Norwegian, Swedish, Icelandic and
Faroese versions of the Universal Dependencies datasets \cite{kromann2004danish,
ovrelid-hohle-2016-universal, nivre-etal-2006-talbanken05,
rognvaldsson2012icelandic, jonsdottir2020creating, arnardottir2020universal}.

Firstly, we assume that the documents in the Universal Dependencies datasets are
grammatically correct, an assumption we have been able to verify for the Danish part,
by manually inspecting a random sample of the documents. We create negative examples by
\textit{either} removing a single word or swapping two consecutive words, where only
one such ``corruption'' is applied to each negative sample.

Naively corrupting the documents in this way does not always lead to grammatically
incorrect samples, however. For instance, removing the word ``rød'' (red) from the
sentence ``Den røde bil er stor'' (The red car is big) does not lead to an incorrect
sentence ``Den bil er stor'' (The car is big).

In order to ensure that the resulting sentence is indeed grammatically correct, we
enforce restrictions on the words that can be removed or swapped. We have gone for a
conservative approach, where we have systematically checked corruptions of words with a
given part-of-speech tag, and only allow corruptions that were always grammatically
correct in our tests. This led us to the following restrictions:

\begin{enumerate}
    \item We do not remove adjectives, adverbs, punctuation, determiners or numbers, as
        the resulting sentence will still be grammatically correct in most cases.
    \item We do not remove nouns or proper nouns if they have another noun or proper
        noun as neighbour, as again that usually does not make the sentence incorrect
        either.
    \item When swapping two neighbouring words, we require them to have different POS
        tags.
    \item We do not swap punctuation or symbols.
    \item If we swap the first word then we ensure that the swapped words have correct
        casing.
\end{enumerate}

We are able to enforce these restrictions as we have gold-standard POS tokens available
for these datasets.

\subsection{Question Answering}
\label{sec:qa}
We also construct new question answering datasets for the Mainland Scandinavian
languages, as we are not aware of any existing datasets for these languages. We dub
these datasets \texttt{ScandiQA}, which we release for each of the Mainland
Scandinavian languages.

These datasets are based on the \texttt{MKQA} dataset \cite{longpre2021mkqa}, which is
based on the \textit{Natural Questions} (\texttt{NQ}) dataset
\cite{kwiatkowski2019natural}. The \texttt{NQ} dataset contains questions inputted to
Google's search engine, associated with the HTML page of the search result. In many
cases these questions have an answer associated with it (a so-called \textit{short
answer}) which appears in the HTML, and in some cases they also have the paragraph in
which the short answer appears (a so-called \textit{long answer}).

The \texttt{MKQA} dataset contains human translations of 10,000 questions and short
answers into 26 languages, including Danish, Norwegian and Swedish. Aside from adding
these translations, the \texttt{MKQA} dataset also corrects many mistakes in the
original \texttt{NQ} dataset by including answers not present in the original dataset,
or by correcting the short answers chosen in the original dataset.

The main thing missing from the \texttt{MKQA} dataset is the context paragraph, which
is what we add to the dataset as follows. For each \texttt{MKQA} sample, we first
locate the corresponding sample in the \texttt{NQ} dataset. If that sample has a long
answer then we use that as the initial (English) context. Otherwise, if neither the
\texttt{NQ} dataset nor the \texttt{MKQA} dataset has an answer registered, then we use
the paragraph in the HTML with the largest cosine similarity to the question, where we
embed the documents using the Sentence Transformer \cite{reimers2019sentence} model
\texttt{all-mpnet-base-v2}.\footnote{\url{https://huggingface.co/sentence-transformers/all-mpnet-base-v2}}

In the last case, where there is no long answer for the sample in \texttt{NQ} but there
\textit{is} an answer in \texttt{MKQA}, we want to identify the paragraph in the HTML
containing the \texttt{MKQA} answer. Unfortunately, the \texttt{MKQA} answers do no
appear verbatim in the HTML (for instance, all dates are standardised to the
\texttt{YYYY-MM-DD} format). We thus start by forming a list of \textit{answer
candidates} based on the \texttt{MKQA} answer, which includes most of the ways dates
and numerals are written in English. We then locate the paragraph containing any of the
answer candidates and which has the largest cosine similarity to the question, where we
embed the documents as described above.

The above procedure thus results in an English context paragraph containing the answer.
We next translate this context paragraph to Danish and Swedish using the DeepL
translation API\footnote{\url{https://www.deepl.com/pro-api}}. As DeepL did not support
Norwegian when we conducted this experiment, we translated the context paragraph to
Norwegian using the Google Translation
API\footnote{\url{https://cloud.google.com/translate/}} instead. With the contexts
translated, we next extract all the answer candidates for the translated context
relevant to the given Mainland Scandinavian language, and change the answer to the
answer candidate appearing in the translated context. If no answer candidate appears in
the translated context then we discard the sample.

The \texttt{MKQA} dataset also contains samples with \textit{no} answer, and we include
these samples in the \texttt{ScandiQA} dataset as well. For these samples, we simply
use the translated context paragraphs as described above. The final dataset contains
7,810 Danish samples, 7,798 Swedish samples and 7,813 Norwegian samples. We release
this dataset separately\footnote{This can be found at \scandiqahfurl\ and the source
code is available at \scandiqarepourl.}, as well as build a ScandEval version of it
with the same train/dev/test size as the other ScandEval datasets. In the ScandEval
version (with 1,024/256/2,048 train/val/test samples as stated in
Section~\ref{sec:uniformdatasets}) we only include samples that contain an answer, as
otherwise we found the 1,024 dataset size to be too small for this task.

We note that since this dataset is a translated version of a dataset originally written
in English, it is not a perfect representation of the Mainland Scandinavian languages,
as many of the questions and answers are concerned with topics specific to the USA.
This might mean that pretrained multilingual models might have an advantage over
monolingual models, but we leave this question for future work.

\subsection{Inference Speed}
\label{sec:speed}
Aside from the predictive performance of the models we also benchmarked the inference
speed of the finetuned models using the \texttt{pyinfer} package
\cite{pyinfer2020github}, and report the mean number of inferences per second. This is
done by recording the mean inference time of running a document with 2,600
characters\footnote{The document is ``This is a dummy document. '', repeated 100
times.} through the model one hundred times, and repeating that process 10 times. We
also compute the confidence interval as described in Section~\ref{sec:bootstrapping}.
These have all been computed using an AMD Ryzen Threadripper 1920X 12-Core CPU.

\section{Benchmarking Package and CLI}
\label{sec:benchmarkingpackageandcli}
To enable every language researcher to benchmark their language models in a
reproducible and consistent manner, we have developed a Python package called
\texttt{scandeval}, which can benchmark any pretrained language model
available on the Hugging Face Hub.

The \texttt{scandeval} package is implemented as both a CLI and a Python package, which
enables ease of use as both a stand-alone benchmarking tool as well as enabling
integration with other Python scripts. The package follows a very \textit{opinionated}
approach to benchmarking, meaning that very few parameters can be changed. This is a
deliberate design decision to enable consistent benchmarking of all models. The package
follows the hyperparameter choices described in Section \ref{sec:finetuningsetup}. See
more in the \texttt{scandeval} documentation.

%
%
%

\section{Experiments}
\label{sec:experiments}
Using the \texttt{scandeval} package we have benchmarked more than 100 pretrained
models in the Scandinavian languages which were available on the Hugging Face Hub.
Aside from these models we also included several multilingual models to enable a fair
comparison. Lastly, to enable better interpretability of the results, we also benchmark
a randomly initialised XLM-RoBERTa-base model \cite{conneau2020unsupervised} and an
ELECTRA-small model \cite{clark2019electra} on the datasets, which will make it more
transparent how much ``external knowledge'' the pretrained models are able to utilise
in their predictions. Benchmarking all these models approximately required 1000 GPU
hours on a GeForce RTX 2080 Ti GPU, which emitted approximately 40 kg of CO$_2$
equivalents\footnote{With a power usage of 250 W/h \cite{tdp} and a carbon efficiency
of 0.16 kg/kWh in Denmark \cite{owidenergy}.}.

\subsection{Benchmarking Results}
\label{sec:benchmarkingresults}
We have presented all of the benchmarked results along with their associated confidence
intervals in an online leaderboard. These
scores have been computed as described in Section~\ref{sec:methodology}, and the top-5
performing models for each language, as well as overall, can be found in
Table~\ref{tab:bestmodels}.

We see from Table~\ref{tab:bestmodels} that
NB-BERT-large\footnote{\url{https://huggingface.co/NbAiLab/nb-bert-large}}
\cite{kummervold-etal-2021-operationalizing} is the best performing model in Norwegian
as well as overall,
DFM-encoder-large-v1\footnote{\url{https://huggingface.co/chcaa/dfm-encoder-large-v1}}
being the best Danish model, and
KB-BERT-large\footnote{\url{https://huggingface.co/KBLab/megatron-bert-large-swedish-cased-165k}}
\cite{malmsten2020swedishbert} having the best performance in Swedish.

The massively multilingual models in the top 5 scores are RemBERT
\cite{chung2020rethinking} and mDeBERTaV3 \cite{he2021debertav3}. The remaining models
in the top 5 are
NB-RoBERTa-base-scandi\footnote{\url{https://huggingface.co/NbAiLab/nb-roberta-base-scandi}},
DanskBERT \cite{snaebjarnarson2023transfer}, NB-BERT-base
\cite{kummervold-etal-2021-operationalizing}, NorBERT2 \cite{kutuzov2021large},
KB-BERT-base \cite{malmsten2020swedishbert} and
AI-Nordics-BERT-large\footnote{\url{https://huggingface.co/AI-Nordics/bert-large-swedish-cased}}.


\begin{table*}[!htb]
    \center
    \scriptsize
    \begin{tabular}{cllll}
        \toprule

        \textbf{Rank} & \textbf{Overall} & \textbf{Danish} & \textbf{Norwegian} &
        \textbf{Swedish} \\

        \midrule

        1 &
        \texttt{NB-BERT-large} &
        \texttt{DFM-encoder-large-v1} &
        \texttt{NB-BERT-large} &
        \texttt{KB-BERT-large} \\

        2 &
        \texttt{DFM-encoder-large-v1} &
        \texttt{NB-BERT-large} &
        \texttt{NB-BERT-base} &
        \texttt{NB-BERT-large} \\

        3 &
        \texttt{RemBERT} &
        \texttt{DanskBERT} &
        \texttt{NB-RoBERTa-base-scandi} &
        \texttt{KB-BERT-base} \\

        4 &
        \texttt{mDeBERTaV3-base} &
        \texttt{RemBERT} &
        \texttt{NorBERT2} &
        \texttt{AI-Nordics-BERT-large} \\

        5 &
        \texttt{NB-RoBERTa-base-scandi} &
        \texttt{mDeBERTaV3-base} &
        \texttt{mDeBERTaV3} &
        \texttt{RemBERT} \\

        \bottomrule

    \end{tabular}
    \caption{The five best performing pretrained models in the Mainland Scandinavian
    language categories.}
    \label{tab:bestmodels}
\end{table*}

\subsection{Cross-lingual Transfer}
\label{sec:crosslingualtransferexperiments}
This experiment investigated the cross-lingual transfer capabilities of the
Scandinavian models, and tested our two hypotheses from Section~\ref{sec:introduction}.
This used the methodology described in Section~\ref{sec:crosslingualtransfer}. For the
Insular Scandinavian languages, the tasks included here are the Icelandic and Faroese
versions of the \texttt{ScaLA} dataset, the Icelandic \texttt{NER} dataset
\texttt{MIM-GOLD-NER} \cite{ingolfsdottir2020named} and the Faroese part of the
\texttt{NER} dataset \texttt{WikiANN} \cite{rahimi-etal-2019-massively}. The resulting
benchmark results can be found in Table~\ref{tab:crosslingualtransfer} and all the raw
scores can be found in the appendix. The results affirm our two hypotheses, as we see
that the group of languages with the largest F-statistic is the group of Mainland
Scandinavian languages.

\begin{table}[!htb]
    \center
    \scriptsize
    \begin{tabular}{lr|lr}

        \toprule
        \textbf{Benchmark group} & \textbf{F-statistic} & \textbf{Benchmark group} &
        \textbf{F-statistic} \\
        \midrule

        \texttt{da}, \texttt{no} & 16.81 &
            \texttt{da}, \texttt{sv}, \texttt{is} & 4.36 \\
        \texttt{da}, \texttt{sv} & 15.48 &
            \texttt{da}, \texttt{sv}, \texttt{fo} & 10.72 \\
        \texttt{da}, \texttt{is} & 4.76 &
            \texttt{da}, \texttt{is}, \texttt{fo} & 5.48 \\
        \texttt{da}, \texttt{fo} & 7.29 &
            \texttt{no}, \texttt{sv}, \texttt{is} & 3.11 \\
        \texttt{no}, \texttt{sv} & 8.14 &
            \texttt{no}, \texttt{sv}, \texttt{fo} & 5.57 \\
        \texttt{no}, \texttt{is} & 3.73 &
            \texttt{no}, \texttt{is}, \texttt{fo} & 3.64 \\
        \texttt{no}, \texttt{fo} & 2.70 &
            \texttt{sv}, \texttt{is}, \texttt{fo} & 4.26 \\
        \texttt{sv}, \texttt{is} & 4.48 &
            \texttt{da}, \texttt{no}, \texttt{sv}, \texttt{is} & 6.97 \\
        \texttt{sv}, \texttt{fo} & 7.59 &
            \texttt{da}, \texttt{no}, \texttt{sv}, \texttt{fo} & 25.40 \\
        \texttt{is}, \texttt{fo} & 21.84 &
            \texttt{da}, \texttt{no}, \texttt{is}, \texttt{fo} & 4.97 \\
        \textbf{\texttt{da}, \texttt{no}, \texttt{sv}} & \textbf{33.34} &
            \texttt{da}, \texttt{sv}, \texttt{is}, \texttt{fo} & 5.21 \\
        \texttt{da}, \texttt{no}, \texttt{is} & 4.27 &
            \texttt{no}, \texttt{sv}, \texttt{is}, \texttt{fo} & 3.38 \\
        \texttt{da}, \texttt{no}, \texttt{fo} & 11.56 &
            \texttt{da}, \texttt{no}, \texttt{sv}, \texttt{is}, \texttt{fo} & 7.53 \\

    \end{tabular}
    \caption{F-statistics showing the cross-lingual transfer between the Scandinavian
    language models. Here \texttt{da} is Danish, \texttt{no} is Norwegian, \texttt{sv}
    is Swedish, \texttt{is} is Icelandic and \texttt{fo} is Faroese.}
    \label{tab:crosslingualtransfer}
\end{table}

\section{Discussion}
\label{sec:discussion}
We note that the benchmarking results presented in
Section~\ref{sec:benchmarkingresults} show that the efforts of the National Libraries
in Norway and Sweden, as well as the Danish Foundation Models project in Denmark, have
paid off, in the sense that their models NB-BERT-large
\cite{kummervold-etal-2021-operationalizing}, KB-BERT-large
\cite{malmsten2020swedishbert} and DFM-encoder-large-v1 are outperforming the
multilingual models.

This seems to indicate that investing in language technologies at a large
language-specific level can be worthwhile. We also see from the same table that the
Norwegian model is within the top two best models in Danish, Norwegian and Swedish,
indicating a potentially large amount of language transfer, supported by the
cross-lingual transfer experiment in Section~\ref{sec:crosslingualtransferexperiments}.
This indicates that a joint Mainland Scandinavian approach could improve the results of
the current monolingual models within the Mainland Scandinavian languages.

\section{Conclusion}
\label{sec:conclusion}
In this paper we have presented a benchmarking framework for the Scandinavian
languages, together with a Python package and CLI, \texttt{scandeval}, which can be
used to benchmark any model available on the Hugging Face Hub. The benchmark features
four tasks: named entity recognition, sentiment classification, linguistic
acceptability and question answering. We have also released two new datasets,
\texttt{ScaLA} and \texttt{ScandiQA}, which constitute the linguistic acceptability and
question answering tasks, respectively. We have benchmarked more than 100 models on the
Mainland Scandinavian datasets in the benchmark and presented these results in an
online leaderboard. In our analysis of the benchmarking results we have shown
substantial cross-lingual transfer between the Mainland Scandinavian languages, and no
notable transfer between the group of Mainland Scandinavian languages and the group of
Insular Scandinavian languages. This is the justification for including only the
Mainland Scandinavian languages in the online leaderboard while maintaining support for
the Insular Scandinavian languages in the \texttt{scandeval} package.

\clearpage
\bibliographystyle{acl_natbib}
\bibliography{nodalida2023}

\clearpage
\appendix
\section{Cross-lingual transfer experiment}
\label{sec:appendixcrosslingualtransfer}

\begin{table}[!htb]
    \begin{tabular}{llc}
        \toprule
        \textbf{Language} & \textbf{Hugging Face Model ID} & \textbf{\# Parameters} \\
        \midrule

        Danish & \texttt{vesteinn/DanskBERT} & 124M \\
        Swedish & \texttt{KB/bert-base-swedish-cased} & 125M \\
        Norwegian & \texttt{patrickvonplaten/norwegian-roberta-base} & 125M \\
        Icelandic & \texttt{mideind/IceBERT} & 124M \\
        English & \texttt{roberta-base} & 125M \\
        German & \texttt{deepset/gbert-base} & 110M \\
        Dutch & \texttt{pdelobelle/robbert-v2-dutch-base} & 117M \\
        Finnish & \texttt{TurkuNLP/bert-base-finnish-cased-v1} & 125M \\
        Russian & \texttt{DeepPavlov/rubert-base-cased} & 178M \\
        Arabic & \texttt{asafaya/bert-base-arabic} & 111M \\

    \end{tabular}
    \parbox{0.9\textwidth}{
        \caption{The Hugging Face Hub model IDs of the models used in the cross-lingual
        transfer experiment.}
    }
    \label{tab:models}
\end{table}

\begin{table}[!htb]
    \footnotesize
    \begin{tabular}{lrrrrr}

        \toprule
        \textbf{Model} & \textbf{Danish Score} & \textbf{Norwegian Score} &
        \textbf{Swedish Score} & \textbf{Icelandic Score} & \textbf{Faroese Score} \\
        \midrule

        \texttt{Danish} &
        \textbf{63.87 ± 1.26} &  
        53.74 ± 3.73 & 
        52.08 ± 2.70 & 
        30.39 ± 1.55 & 
        45.26 ± 1.35 \\ 

        \texttt{Norwegian} &
        46.30 ± 2.83 & 
        \textbf{58.78 ± 1.44} & 
        46.90 ± 2.79 & 
        28.85 ± 1.45 & 
        43.35 ± 2.20 \\ 

        \texttt{Swedish} &
        45.81 ± 2.96 & 
        47.32 ± 2.66 & 
        \textbf{69.29 ± 1.40} & 
        28.69 ± 1.61 & 
        43.63 ± 1.90 \\ 

        \texttt{Icelandic} &
        30.20 ± 1.23 & 
        28.68 ± 2.91 & 
        36.80 ± 2.14 & 
        \textbf{71.00 ± 1.50} & 
        \textbf{48.26 ± 4.76} \\ 

        \texttt{Finnish} &
        32.55 ± 1.47 & 
        30.71 ± 2.14 & 
        38.94 ± 1.51 & 
        16.33 ± 1.89 & 
        36.87 ± 1.17 \\ 

        \texttt{English} &
        34.11 ± 2.11 & 
        30.92 ± 2.69 & 
        39.24 ± 1.92 & 
        28.39 ± 2.41 & 
        40.75 ± 1.59 \\ 

        \texttt{German} &
        28.13 ± 2.04 & 
        27.58 ± 2.90 & 
        37.62 ± 4.18 & 
        26.13 ± 1.63 & 
        41.02 ± 1.46 \\ 

        \texttt{Dutch} &
        31.78 ± 1.62 & 
        28.27 ± 2.51 & 
        35.06 ± 1.87 & 
        26.21 ± 1.79 & 
        40.83 ± 1.70 \\ 

        \texttt{Russian} &
        33.91 ± 1.88 & 
        33.55 ± 2.17 & 
        39.14 ± 2.33 & 
        29.96 ± 1.58 & 
        43.17 ± 1.66 \\ 

        \texttt{Arabic} &
        22.89 ± 1.82 & 
        19.98 ± 2.24 & 
        25.40 ± 2.69 & 
        10.33 ± 2.19 & 
        35.33 ± 1.57 \\ 

    \end{tabular}
    \parbox{0.9\textwidth}{
        \caption{The raw benchmarking results used in the cross-lingual transfer
        experiment.}
    }
    \label{tab:rawcrosslingualtransfer}
\end{table}

\clearpage
\onecolumn
\section{Training Data Size Experiment}
\label{sec:appendixtrainingsize}

\begin{figure}[!htb]
    \includegraphics[width=1.0\textwidth]{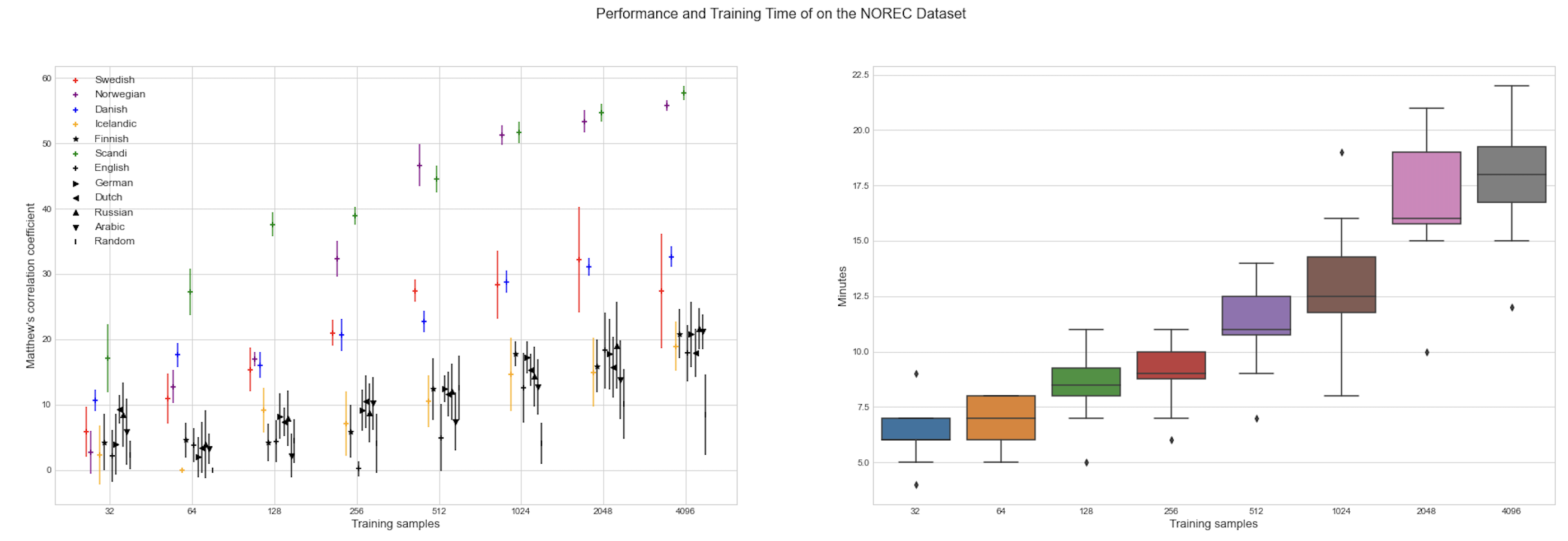}
    \caption{The results from the training data size experiment for the \texttt{NoReC}
    dataset.}
    \label{fig:norectrainingsize}
\end{figure}

\begin{figure}[!htb]
    \includegraphics[width=1.0\textwidth]{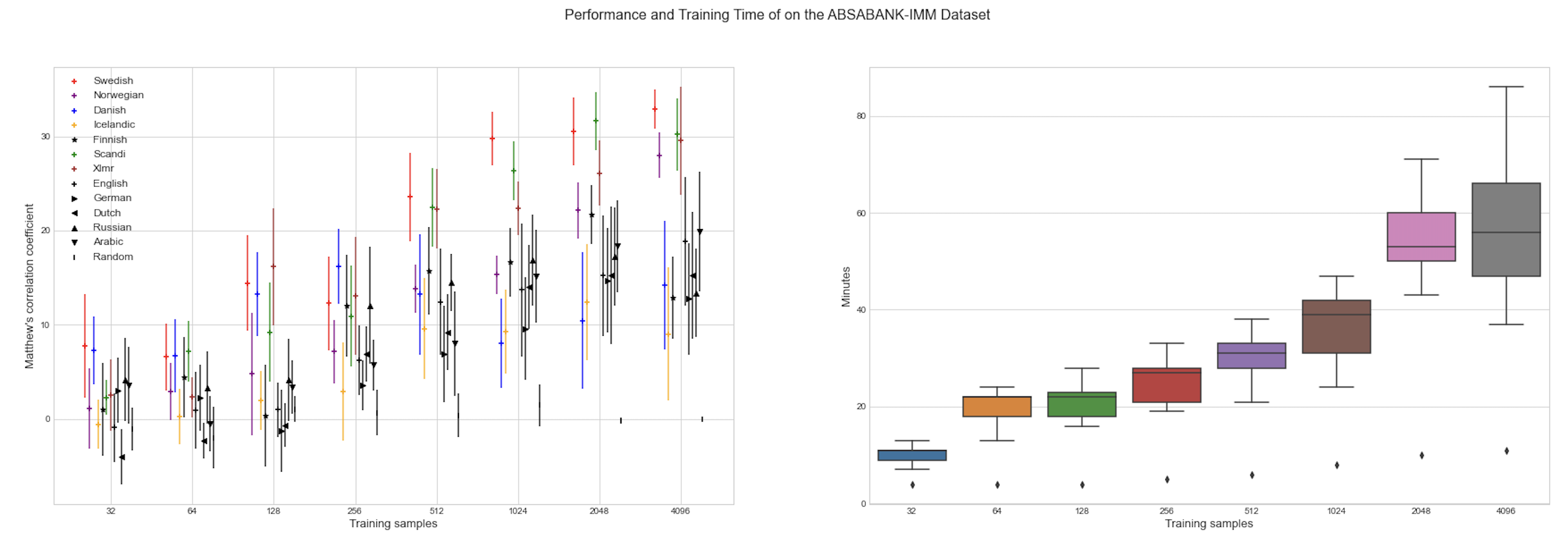}
    \caption{The results from the training data size experiment for the
    \texttt{Absabank-Imm} dataset \cite{adesam2020swedishglue}.}
    \label{fig:absabankimmtrainingsize}
\end{figure}

\begin{figure}[!htb]
    \includegraphics[width=1.0\textwidth]{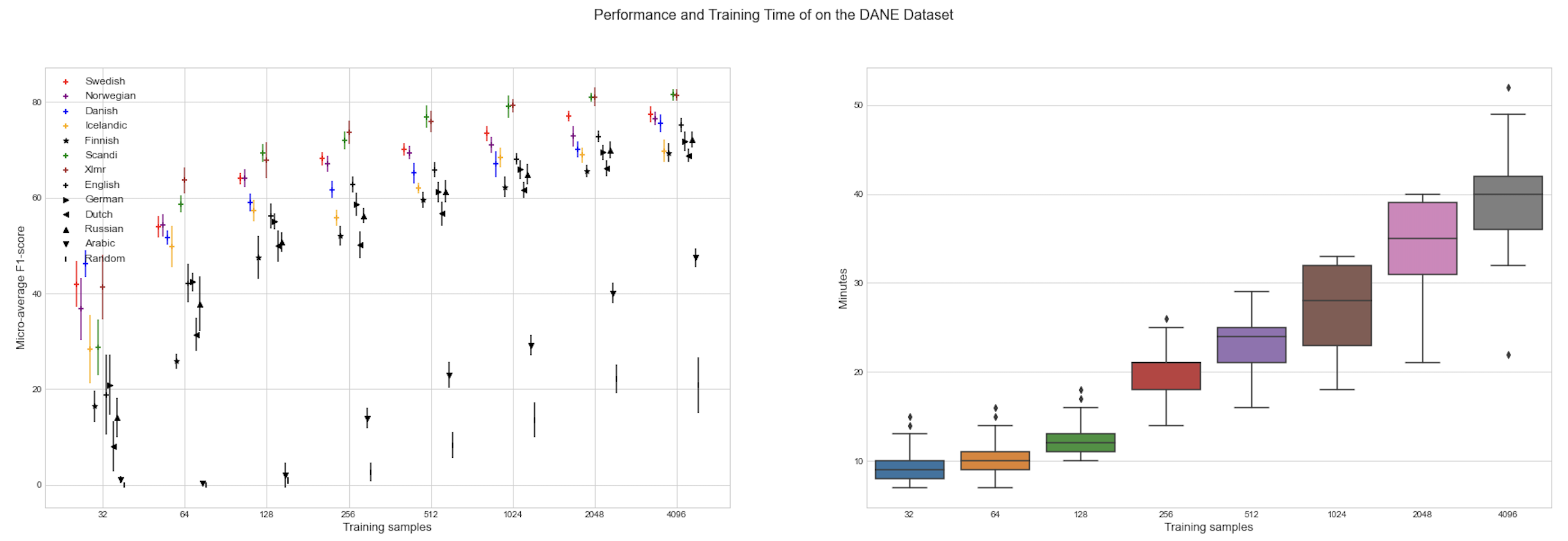}
    \caption{The results from the training data size experiment for the
        \texttt{DaNE} dataset.}
    \label{fig:danetrainingsize}
\end{figure}

\begin{figure}[!htb]
    \includegraphics[width=1.0\textwidth]{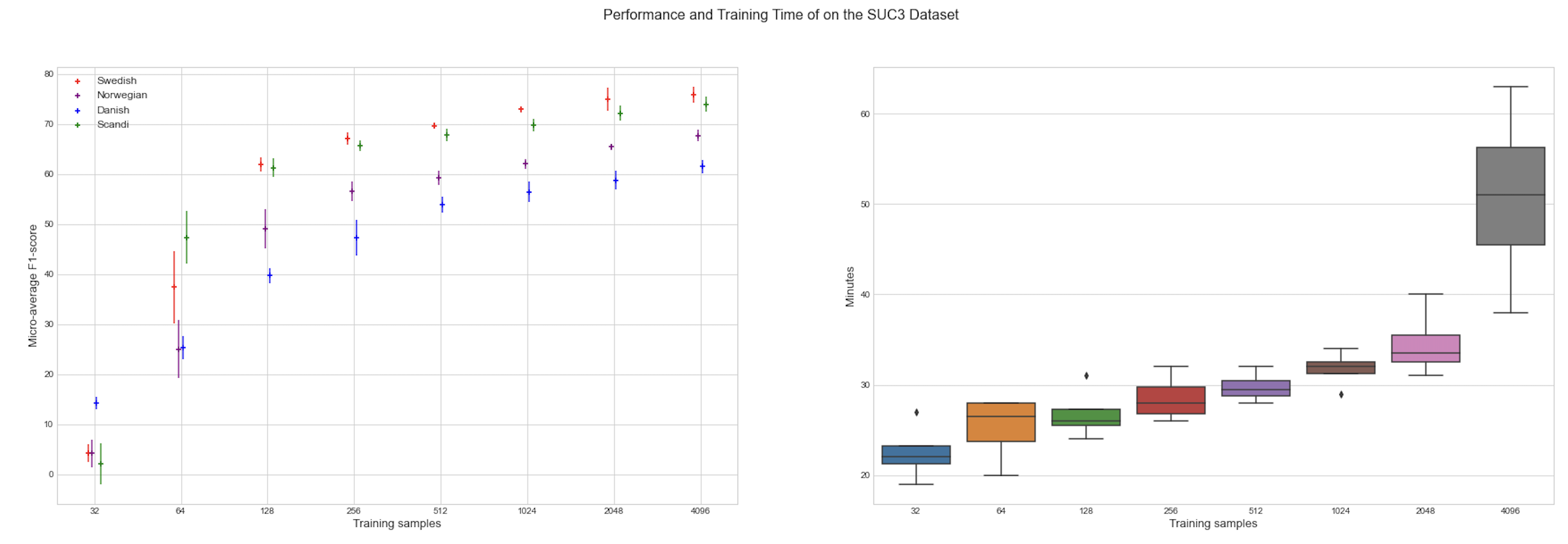}
    \caption{The results from the training data size experiment for the
        \texttt{SUC3} dataset.}
    \label{fig:suc3trainingsize}
\end{figure}

\begin{figure}[!htb]
    \includegraphics[width=1.0\textwidth]{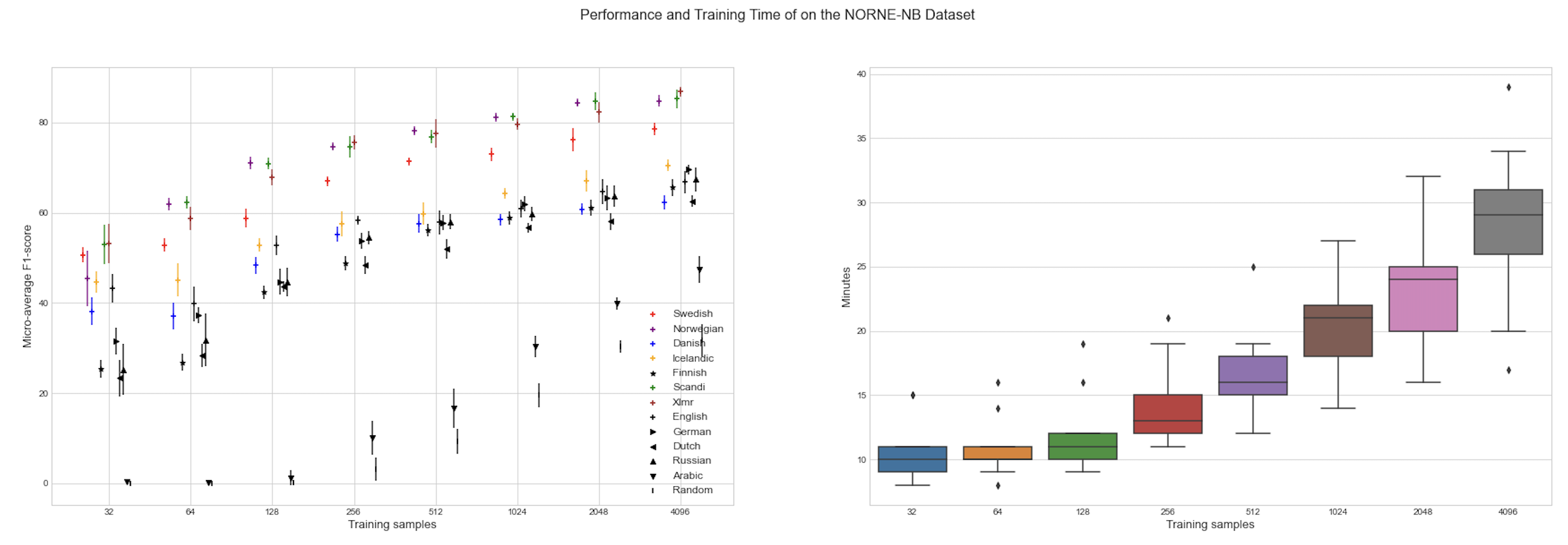}
    \caption{The results from the training data size experiment for the
        \texttt{NorNE-NB} dataset.}
    \label{fig:nornenbtrainingsize}
\end{figure}

\begin{figure}[!htb]
    \includegraphics[width=1.0\textwidth]{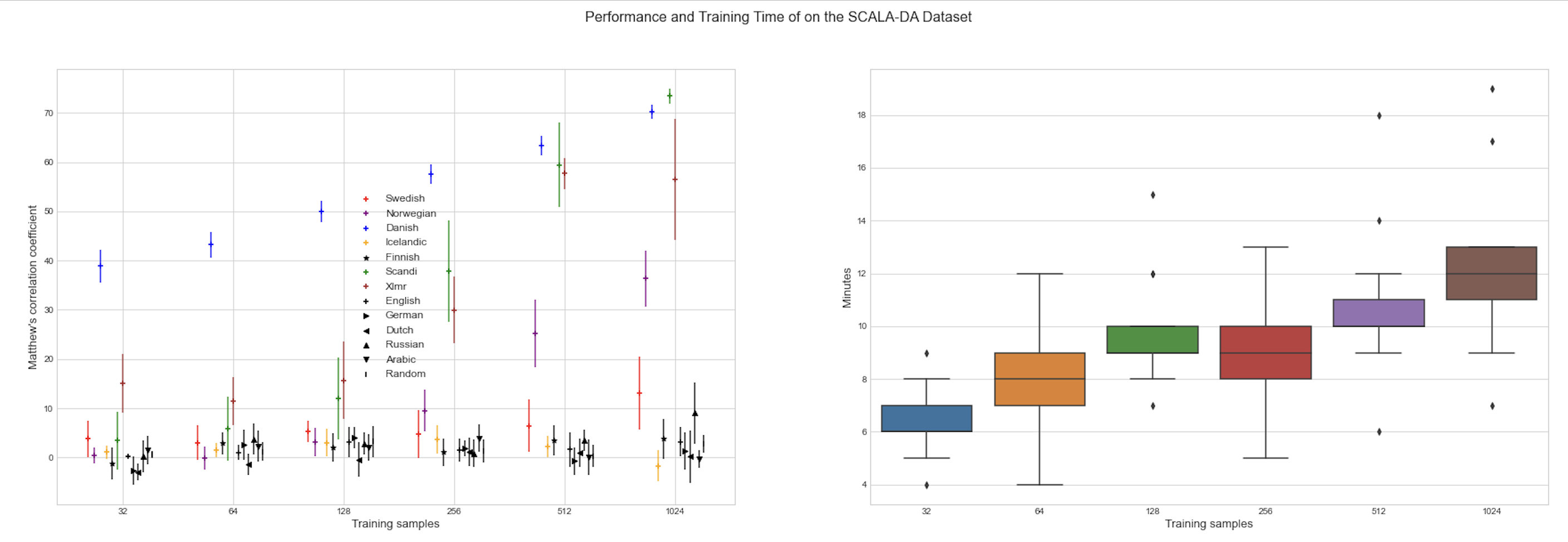}
    \caption{The results from the training data size experiment for the
        \texttt{ScaLA-DA} dataset.}
    \label{fig:scaladatrainingsize}
\end{figure}

\begin{figure}[!htb]
    \includegraphics[width=1.0\textwidth]{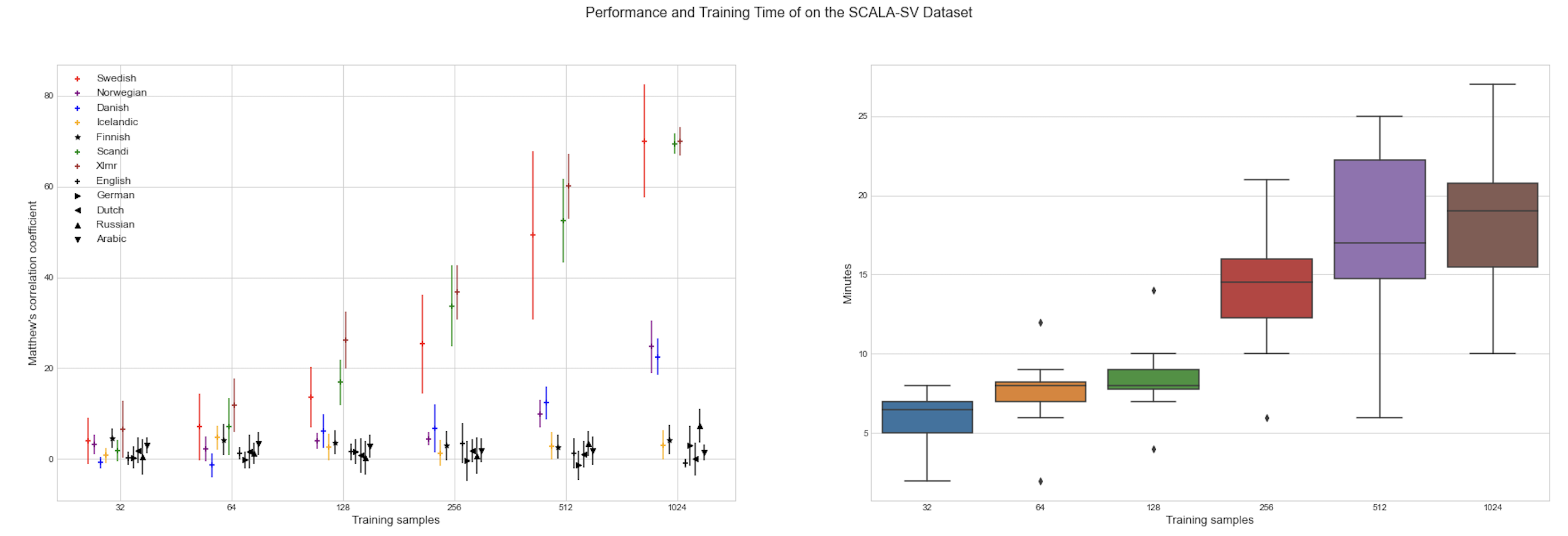}
    \caption{The results from the training data size experiment for the
        \texttt{ScaLA-SV} dataset.}
    \label{fig:scalasvtrainingsize}
\end{figure}

\begin{figure}[!htb]
    \includegraphics[width=1.0\textwidth]{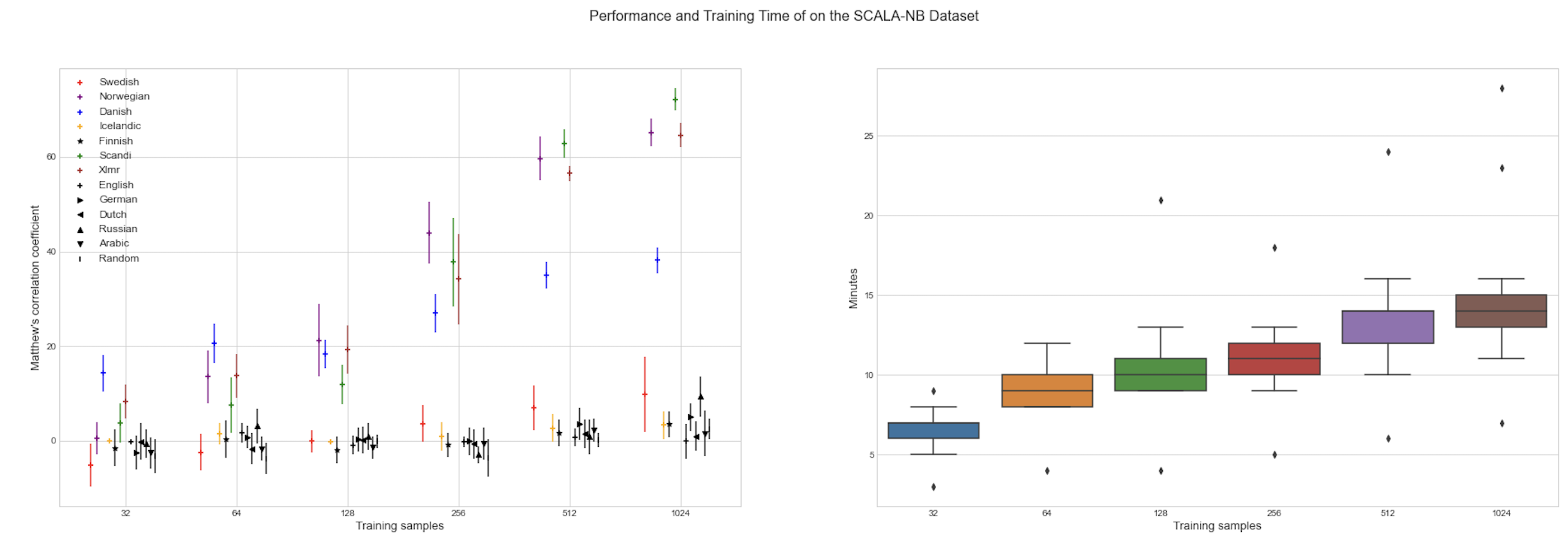}
    \caption{The results from the training data size experiment for the
        \texttt{ScaLA-NB} dataset.}
    \label{fig:scalanbtrainingsize}
\end{figure}

\begin{figure}[!htb]
    \includegraphics[width=1.0\textwidth]{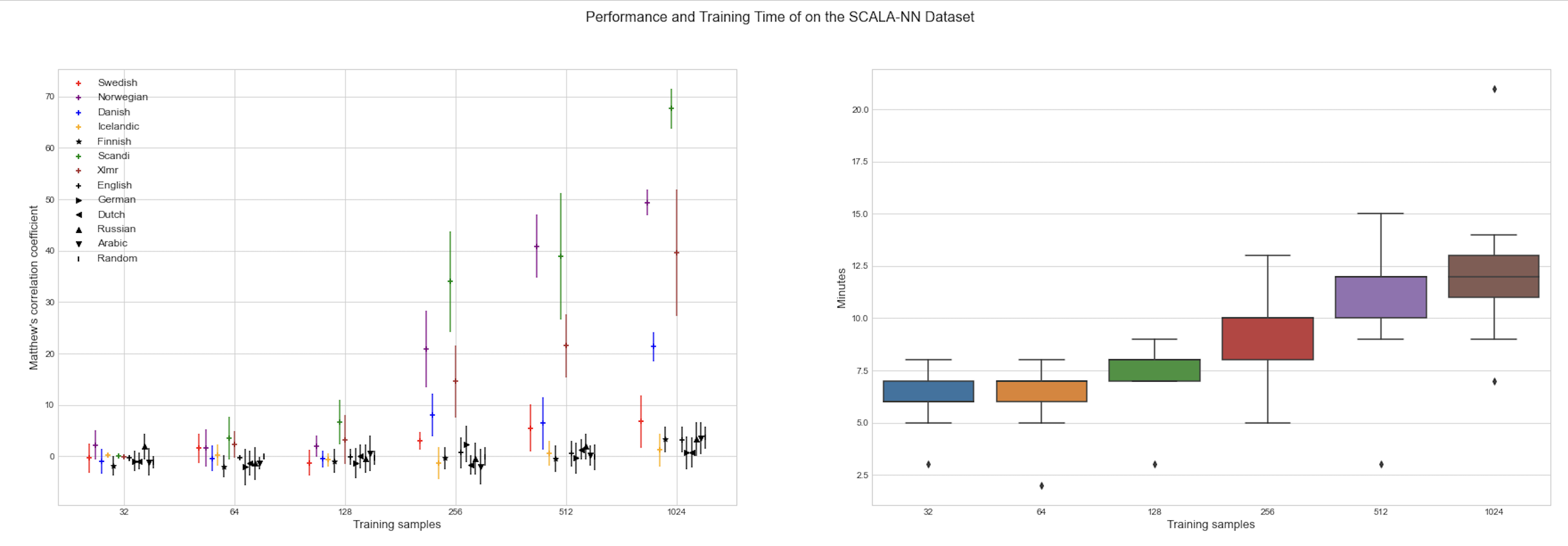}
    \caption{The results from the training data size experiment for the
        \texttt{ScaLA-NN} dataset.}
    \label{fig:scalannttrainingsize}
\end{figure}

\end{document}